\RequirePackage{amsthm}

\documentclass[sn-mathphys-num]{sn-jnl}

\usepackage{graphicx}
\usepackage{multirow}
\usepackage{amsmath,amssymb,amsfonts}
\usepackage{amsthm}
\usepackage{mathrsfs}
\usepackage[title]{appendix}
\usepackage{xcolor}
\usepackage{textcomp}
\usepackage{manyfoot}
\usepackage{booktabs}
\usepackage{algorithm}
\usepackage{algorithmicx}
\usepackage{algpseudocode}
\usepackage{listings}
\usepackage{lmodern}

\theoremstyle{thmstyleone}

\theoremstyle{thmstyletwo}

\theoremstyle{thmstylethree}

\begin{document}

\title[Enhanced Interactive-Voting Based Map Matching]{Enhancing Interactive Voting-Based Map Matching: Improving Efficiency and Robustness for Heterogeneous GPS Trajectories}

\author*[1]{\fnm{William} \sur{Alemanni}}\email{will.alemanni2@gmail.com}
\author[1,2]{\fnm{Arianna} \sur{Burzacchi}}
\author[1]{\fnm{Davide} \sur{Colombi}}
\author[1]{\fnm{Elena} \sur{Giarratano}}

\affil[1]{\orgname{Cuebiq Group, LLC}, \orgaddress{\city{Milan}, \country{Italy}}}
\affil[2]{\orgdiv{MOX Laboratory}, \orgname{Department of Mathematics, Politecnico di Milano}, \orgaddress{\city{Milano}, \country{Italy}}}

\abstract{
This paper presents an enhanced version of the Interactive Voting-Based Map Matching algorithm, designed to efficiently process trajectories with varying sampling rates. 
The main aim is to reconstruct GPS trajectories with high accuracy, independent of input data quality. 
Building upon the original algorithm, developed exclusively for aligning GPS signals to road networks—we extend its capabilities by integrating trajectory imputation. 

Our improvements also include the implementation of a distance-bounded interactive voting strategy to reduce computational complexity, as well as modifications to address missing data in the road network. 
Furthermore, we incorporate a custom-built asset derived from OpenStreetMap, enabling this approach to be smoothly applied in any geographic region covered by OpenStreetMap’s road network. 
These advancements preserve the core strengths of the original algorithm while significantly extending its applicability to diverse real-world scenarios
}


\maketitle

\section{Introduction}\label{sec1}

In recent years, location acquisition technologies have seen dramatic growth, with the Global Positioning System (GPS) generating extensive trajectory data for a variety of applications. 
However, raw GPS trajectories are often affected by noise, low sampling frequencies, and non-uniform intervals. 
These issues arise from factors such as urban signal obstructions, energy and bandwidth constraints, device inconsistencies, and increasing privacy concerns \citep{dogramadzi2021accelerated, andres2013geo, haydari2022differentially}. 
As a result, advanced techniques are required to ensure accurate and reliable trajectory analysis.

Map matching, the process of aligning GPS trajectories with the underlying road network, addresses these challenges by associating noisy GPS points with the most likely road segments. 
This step is essential in geographic information systems (GIS) and location-based services, supporting critical applications such as traffic flow analysis, travel time estimation, trajectory enhancement, and navigation services \citep{xu2019utilizing, xu2022intelligent, li2021trajectory, jiang2023l2mm}.

Despite notable advances in map matching algorithms, accurately matching trajectories remains challenging, especially with low-sampling-rate data characterized by significant gaps between consecutive points \citep{lou2009map}. 
Traditional methods, typically designed for high-frequency data, often struggle under these conditions, leading to increased uncertainty in reconstructing plausible paths. 
Furthermore, the complexity of urban road networks, characterised by elements such as parallel roads and multi-level intersections, may creates ambiguity in the matching process \citep{fang2022map, huang2021survey}.

Map matching techniques are generally classified into three categories: global, local, and probabilistic \citep{quddus2007current, yuan2010interactive}. 
Global methods evaluate entire trajectories to determine the optimal matching path, often employing distance measures such as the Fréchet distance \citep{brakatsoulas2005map}. 
Although effective for low-frequency data, these methods tend to be computationally intensive. 

Local methods, such as point-to-point and point-to-curve matching, sequentially align individual GPS points with nearby road segments based on spatial proximity and network topology \citep{lou2009map}. 
While these are computationally efficient, they generally underperform when faced with sparse data or complex road networks. 

In contrast, probabilistic approaches—especially those based on Hidden Markov Models (HMM)—model the problem as a sequence of hidden states (true positions) and observations (GPS points), using emission and transition probabilities to infer the most likely path \citep{newson2009hidden, jagadeesh2017probabilistic, song2018hidden}.

Recent developments in machine learning and deep learning have shown promise in addressing the complexities inherent in map matching by capturing intricate spatial and temporal dependencies \citep{jin2022transformer, fang2022map}. 
However, these models can be difficult to interpret and fine-tune, limiting their broad applicability in real-world scenarios.

In this study, we enhance the Interactive Voting-based Map Matching (IVMM) algorithm originally proposed by \citet{yuan2010interactive}. 
IVMM integrates local, global, and probabilistic strategies into a flexible framework, making it particularly effective for low-sampling-rate GPS data.
By employing a weighted mutual influence approach alongside a distance-based weighting mechanism, the algorithm efficiently captures spatial relationships and enhances matching accuracy in complex scenarios. 
Unlike machine learning models, IVMM maintains transparency and offers a balanced approach to handle different use cases.

Our primary goal is to reconstruct GPS trajectories with high accuracy, regardless of the quality of the input data.
This objective encompasses scenarios ranging from high-quality, high-frequency data to datasets with gaps and lower sampling rates.
The IVMM algorithm was originally designed only for map matching—aligning GPS pings to road networks. In this work, we extend its capabilities by integrating trajectory imputation \cite{zheng2012reducing, long2016kinematic, li2021trajectory}, which allows us to infer missing segments between paired points using the same interactive voting method.
To address sparse GPS data, we leverage this methodology to reconstruct missing paths by choosing the shortest road network path between consecutive matched candidates.
This improvement does not involve any additional computational overhead, as it exploits the same voting structure for both map matching and trajectory imputation.
We introduce several key enhancements to the original IVMM algorithm:
\begin{itemize}
    \item A distance-bounded voting implementation that reduces computational complexity, making the approach suitable for large-scale applications.
    \item Techniques for handling missing paths in road networks, thereby improving robustness in scenarios with incomplete road network data.
    \item Adaptation for use with OpenStreetMap \citep{OpenStreetMap}, broadening the algorithm’s applicability.
    \item Trajectory imputation via the interactive voting method, which reconstructs missing segments by inferring plausible paths between matched GPS points.
\end{itemize}

We evaluate the performance of the enhanced IVMM algorithm using two distinct datasets: connected car data (representing high-sampling-rate trajectories) and mobile device location data (representing low-sampling-rate trajectories). 
Both datasets are integrated with road network data from OpenStreetMap. 
Our experimental results indicate improvements in the computational efficiency and applicability of the model.

The remainder of this paper is organized as follows. 
We first present the experimental results, highlighting the performance gains achieved by the enhanced IVMM algorithm. 
Next, we discuss the implications of our findings in the broader context of location-based services and trajectory analysis, and suggest directions for future research. 
Finally, we provide a detailed description of our methodology, including data sources and the specific enhancements made to the IVMM algorithm, to facilitate reproducibility and further investigation by the research community.
\section{Results}\label{sec2}

In this section, we present the experiments conducted and the results obtained from implementing the ping map matching algorithm.
We do not evaluate the algorithm's performance in terms of matching accuracy, as its effectiveness has already been established and compared with other state-of-the-art methods in \citet{yuan2010interactive}.
Instead, our focus is on evaluating the improvements in algorithm applicability achieved through our enhancements.

First, we describe the metrics used, followed by a presentation of the results and key findings.
Since ground truth data is unavailable for all datasets used in our experiments, the exact location of each ping remains unknown.
As a result, conventional validation methods—typically based on comparing matched points with actual device locations—cannot be applied.
To address this limitation, we defined three metrics (Table \ref{tab:metrics}) to assess result quality without relying on a reference path.
The only performance metric we consider is computational time, which we use to measure the improvement of the distance-bounded model compared to the version by Yuan et al. 
Additionally, two descriptive metrics are introduced: the Ping-candidate distance and the Path length variation. 
These are used to describe the results and, in particular, to control that the results remain stable even when the computational-wise improvements we have introduced are applied.

\begin{table}[t]
    \centering
    \renewcommand{\arraystretch}{1}
    \begin{tabular}{l p{5.5cm} c}
        \toprule
        \textbf{Metric} & \textbf{Description} & \textbf{Metric type}\\
        \midrule
        Computational time (seconds) & Time taken to process the input data and perform the method & Performance\\
        Ping-candidate distance (meters) & Distances between pings and their corresponding candidate on the road network & Descriptive\\
        Path length variation (meters) & Difference in the length of the trajectory, where lengths are computed as cumulative distance of consecutive pings or consecutive candidates on the road network and then compared & Descriptive\\
        \bottomrule
    \end{tabular}
    \caption{Metrics table}
    \label{tab:metrics}
\end{table}

\begin{figure}[t!]
    \centering
    \includegraphics[width=\linewidth]{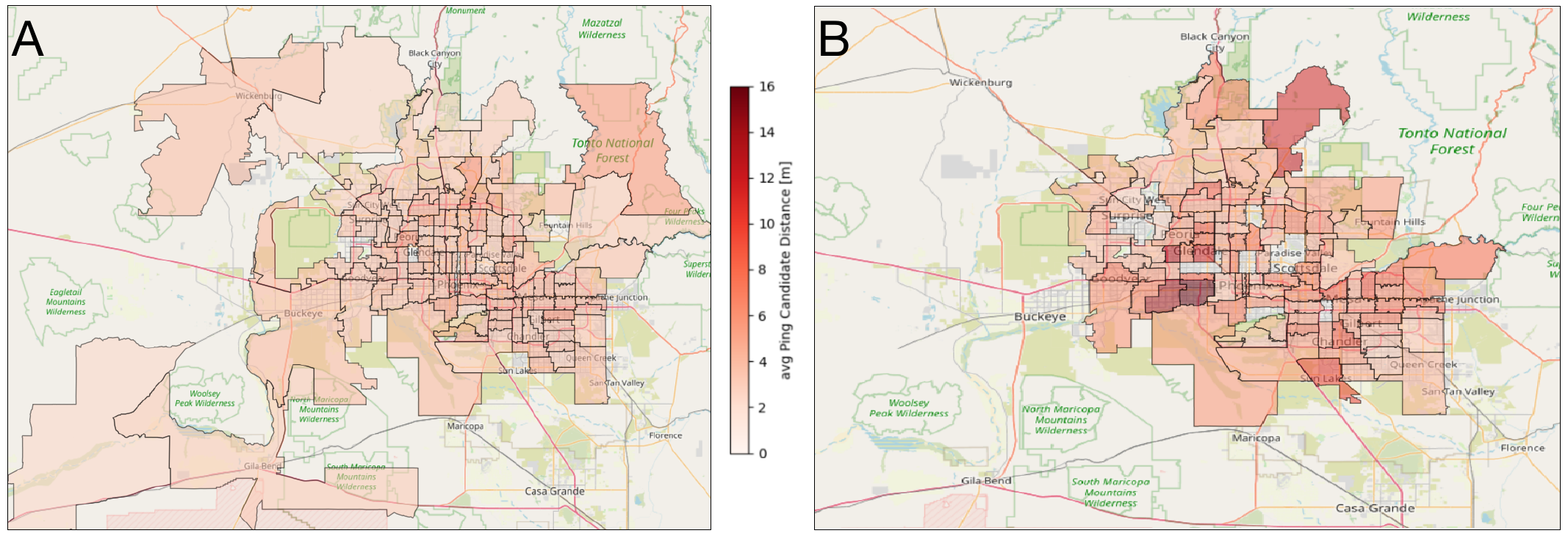}
    \caption{\textbf{Ping-candidate distance for vehicles and mobile devices}. 
	\textbf{A), B)} The average distance between the original GPS point (ping) and the best candidate on the road network obtained through the algorithm and aggregated at the zip-code level respectively for vehicles data and mobile device data. 
	}
    \label{fig:results_map}
\end{figure}

\begin{figure}[t!]
    \centering
    \includegraphics[width=\linewidth]{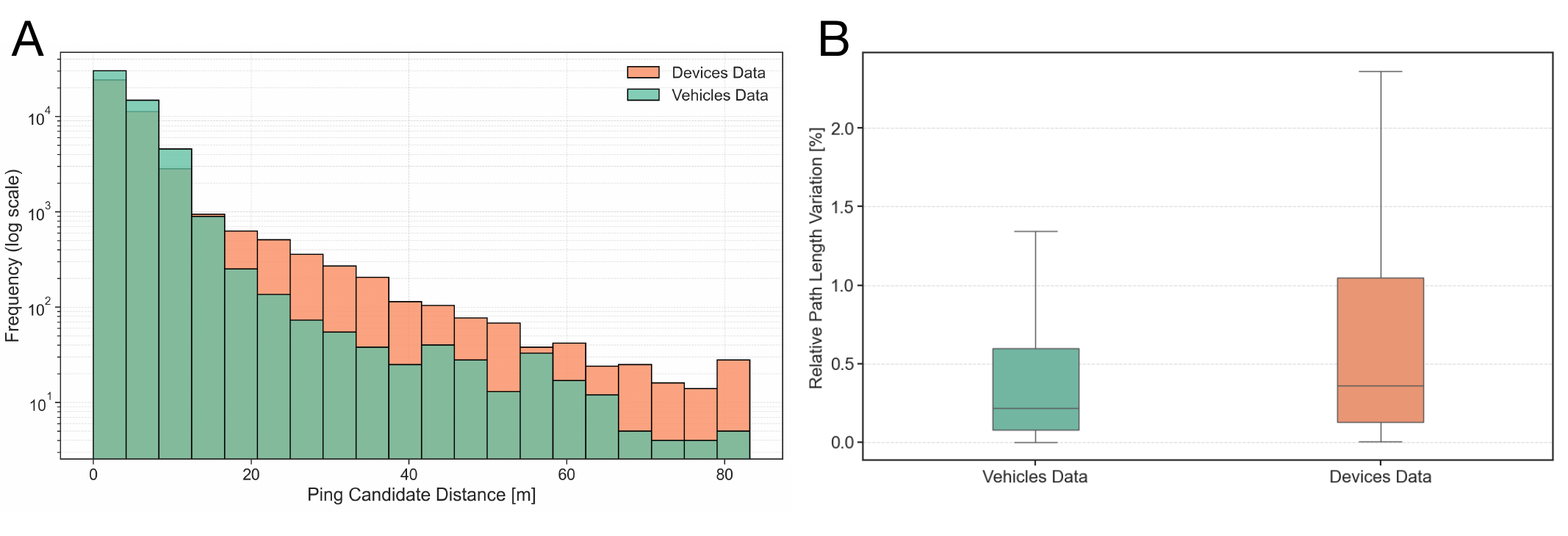} 
    \caption{\textbf{Descriptive metrics behaviour}. 
	\textbf{A)} Distribution of the ping candidate distance for vehicles data (\textit{green}) and mobile phone data (\textit{orange}).
	\textbf{B)} Distribution of the relative path lenght variation for vehicles data (\textit{green}) and mobile phone data (\textit{orange}).
	}
    \label{fig:results_metrics}
\end{figure}

The study utilizes two separate anonymized and privacy-enhanced GPS trajectory datasets obtained from connected vehicles and mobile phones in Maricopa County, Arizona, USA.

Figure \ref{fig:results_map} highlights two key observations.
First, the connected vehicle dataset (Fig \ref{fig:results_map} A) extends beyond the urban area of Maricopa County, encompassing zip codes outside the city and including GPS points along long-distance roads, such as highways.

In contrast, the mobile phone dataset (Fig \ref{fig:results_map} B) exhibits a higher concentration of GPS points within the urban area.
This distinction is particularly relevant because, as described in the methods section, GPS signals tend to be less accurate in urban environments.
Additionally, the increased complexity of the road network in urban areas makes the map-matching process more challenging.

For instance, in both datasets, zip codes with the highest average ping candidate distance are those located in the city's central areas.
As expected, mobile phone GPS data tends to be less accurate and features a more heterogeneous sampling rate, typically resulting in higher ping candidate distances.

This trend is further illustrated in the histogram in Fig \ref{fig:results_metrics} A, which shows the distribution of ping candidate distances for both datasets.
Figure \ref{fig:results_metrics} B presents the relative percentage difference between the length of the original and matched paths (as defined in Table \ref{tab:metrics}).
For both datasets, this difference remains generally low, with a median below 0.5\%.
However, the distribution for mobile phone data is skewed toward higher percentage values, reflecting an increased variability in ping-candidate distance.
This implies that for mobile phone data, the matching strategy must deviate more from simple ping-nearest-candidate matching to ensure a consistent and reasonable trajectory.

\subsection{Distance-Bounded Interactive Voting}

One of the main challenges in applying the interactive voting approach to large-scale datasets is its high computational cost, which increases quadratically with the number of pings and candidates.
To mitigate this issue, we implemented a version of the voting process that limits interactions to pings within a predefined distance threshold. 
This optimization is possible because, in the voting process of the algorithm proposed by \citet{yuan2010interactive}, each candidate for a given ping contributes to the selection of optimal candidates for all other pings, with an influence that decreases as distance increases (see \ref{algo_description}). 

However, at the application level, only nearby candidates have a significant impact. 
To accelerate execution, we therefore limited the voting process to candidates within a meaningful distance, excluding those with negligible influence.
This modification significantly reduces computational complexity while preserving the key characteristics of the original method.

In this section, we analyze the impact of this distance-bounded implementation by evaluating its performance across different distance thresholds.
Our results focus on the trade-off between processing speed and map-matching performances, providing insights into a good balance between efficiency and precision.
For the same GPS sample, in this case we considered only vehicles data, we run the algorithm using the original voting method (in the following named \textit{global implementation}) and the distance-bounded implementation method, 
with varying distance threshold (\textit{maxdist}) equal to 1000, 2500, 4000, and 5500 meters. 

\begin{figure}[t!]
    \centering
    \includegraphics[width=0.7\linewidth]{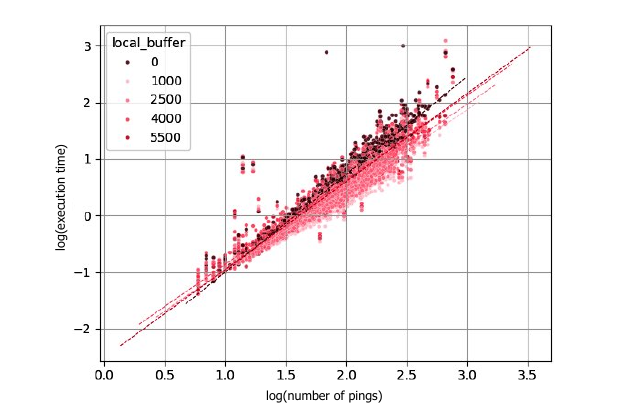} 
    \caption{\textbf{Execution time analysis}.  
    Relationship between execution time and key factors in logarithmic scale.  
    Execution time as a function of the log of the number of pings, with different \textit{distance buffer} values in meters.
    }
    \label{fig:execution_time}
\end{figure}

\begin{table}
    \centering
    \renewcommand{\arraystretch}{1.1}
    \resizebox{\textwidth}{!}{
        \begin{tabular}{ |c | c | c | c | c | c | c | c | c |}
            \hline
            & Type & avg (sec) & std (sec) & min (sec) & q1 (sec) & q2 (sec) & q3 (sec) & max (sec) \\
            \hline
            \multirow{2}{*}{Global} & Full Process & \textbf{13.3} & 52.5 & 0.07 & 0.9 & \textbf{3.0} & 10.1 & 1006.0 \\
            & Voting step & \textbf{12.2} & 50.9 & 0.06 & 0.8 & \textbf{2.7} & 8.9 & 994.5 \\
            \hline
            \multirow{2}{*}{5500 m} & Full Process & \textbf{8.3} & 31.3 & 0.07 & 0.8 & \textbf{2.6} & 7.4 & 831.1 \\
            & Voting step & \textbf{7.2} & 29.5 & 0.05 & 0.7 & \textbf{2.3} & 6.3 & 796.1 \\
            \hline
            \multirow{2}{*}{4000 m} & Full Process & \textbf{8.7} & 28.7 & 0.06 & 1.0 & \textbf{2.9} & 7.6 & 663.4 \\
            & Voting step & \textbf{7.5} & 26.7 & 0.04 & 0.8 & \textbf{2.5} & 6.5 & 637.6 \\
            \hline
            \multirow{2}{*}{2500 m} & Full Process & \textbf{7.3} & 41.8 & 0.06 & 0.7 & \textbf{2.1} & 5.8 & 1245.9 \\
            & Voting step & \textbf{6.3} & 40.7 & 0.04 & 0.7 & \textbf{1.8} & 4.6 & 1223.9 \\
            \hline
            \multirow{2}{*}{1000 m} & Full Process & \textbf{7.1} & 32.8 & 0.06 & 0.7 & \textbf{1.8} & 5.2 & 852.4 \\
            & Voting step & \textbf{5.8} & 30.3 & 0.04 & 0.6 & \textbf{1.5} & 4.1 & 826.0 \\
            \hline
        \end{tabular}
    }
    \caption{Summary statistics of the computational time comparing full procedure vs voting procedure}
    \label{tab:time_variations}
\end{table}

Table \ref{tab:time_variations} presents general statistics on the processing times of approximately 500 trajectory runs. 
Specifically, we distinguish between the time required to complete the entire process (\textit{Full Process}) and the time spent on the most computationally expensive step (\textit{Voting step}), comparing both the original algorithm and our implementation with different threshold values.

The results indicate that the voting phase consistently represents the most time-consuming component across all executions. 
Moreover, as the \textit{maxdist} buffer width decreases, the computational time is reduced in both mean and median values. 
This trend is accompanied by a lower variability in computational time relative to the mean, as reflected in a smaller standard deviation.

To test the different relationships between computational time and the number of pings for the different implementations analyzed, in Fig \ref{fig:results_metrics}, a linear regression model has been used ($R^2 = 0.91$).
The slope of the regression line is higher for the implementation with the widest \textit{maxdist} and the global implementation and lower for lower \textit{maxdist}; 
this means that especially with large numbers of pings, the distance-bounded implementation can decrease significatively the computational time.

\begin{table}[t!]
    \centering
    \renewcommand{\arraystretch}{1.1}
    \resizebox{\textwidth}{!}{
        \begin{tabular}{|c|ccccc|ccccc|}
            \hline
            & \multicolumn{5}{c|}{\textbf{Candidate-ping distance [m]}} & \multicolumn{5}{c|}{\textbf{Trajectory length variation [m]}} \\
            \hline
            \textbf{Buffer Size} & \textbf{min} & \textbf{q1} & \textbf{q2} & \textbf{q3} & \textbf{max} & \textbf{min} & \textbf{q1} & \textbf{q2} & \textbf{q3} & \textbf{max} \\
            \hline
            1000 & 0.265 & 1.950 & \textbf{3.753} & 6.141 & 18.061 & 12.688 & 24.193 & \textbf{262.555} & 20.197 & 26.519 \\
            2500 & 0.265 & 1.951 & \textbf{3.754} & 6.142 & 18.072 & 12.646 & 24.235 & \textbf{263.307} & 20.175 & 26.466 \\
            4000 & 0.265 & 1.951 & \textbf{3.754} & 6.143 & 18.076 & 12.646 & 24.193 & \textbf{260.974} & 20.155 & 26.407 \\
            5500 & 0.265 & 1.951 & \textbf{3.754} & 6.143 & 18.074 & 12.646 & 24.235 & \textbf{259.132} & 20.166 & 26.398 \\
            Global & 0.265 & 1.950 & \textbf{3.754} & 6.142 & 18.058 & 12.623 & 24.235 & \textbf{286.648} & 20.251 & 26.885 \\
            \hline
        \end{tabular}
    }
    \caption{Combined summary statistics of the candidate-ping distance and trajectory length variation.}
    \label{tab:combined_summary_stats}
\end{table}

After confirming that the computational time decreases-as indicator of performance improving- as the threshold set by \textit{maxdist} decreases, we present in Table \ref{tab:combined_summary_stats} the summary statistics for the two metrics we considered to monitor quality stability: Candidate-ping distance and Trajectory length variation.
Regarding the Candidate-ping distance, all implementations yield identical values in terms of summary statistics confirming that the goodness of the model has been preserved.
A similar pattern is observed for the Trajectory length variation, where the results differ only slightly across implementations, with variations of at most a few tens of meters.

\subsection{Handling Missing Paths}

\begin{figure}[ht!]
    \centering
    \includegraphics[width=1.\linewidth]{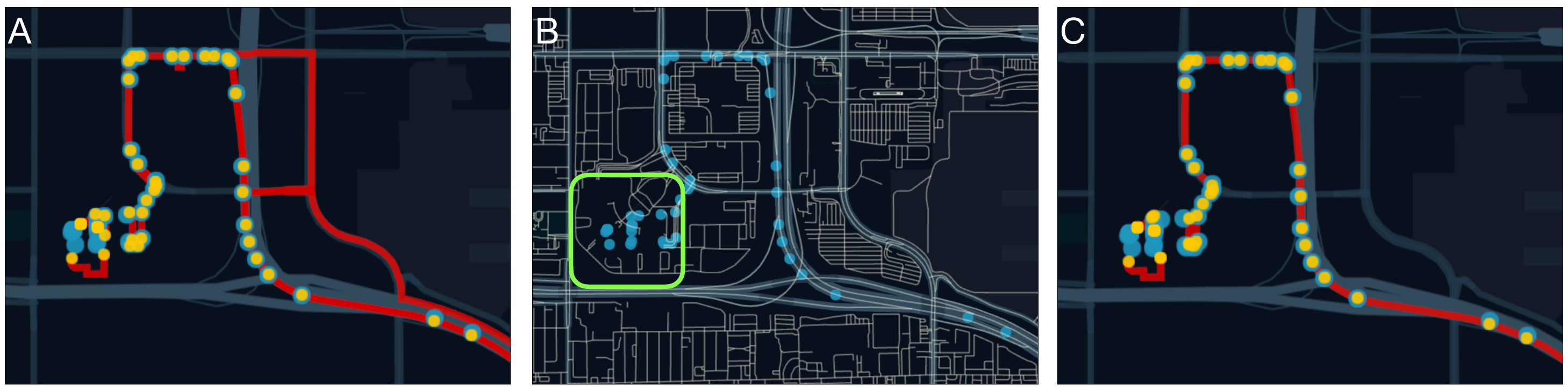} 
    \caption{\textbf{Missing path example}.  
    \textbf{A)} The matched path (in \textit{red}) obtained by the best candidates (in \textit{yellow}) given by the trajectory points (in \textit{light blue}) using the \citet{yuan2010interactive} approach.
    \textbf{B)} The road pieces network (in \textit{white}) in the area where the trajectory is located. In \textit{green} is highlighted the part where there are missing road pieces.
    \textbf{C)} The matched path (in \textit{red}) obtained by the best candidates (in \textit{yellow}) given by the trajectory points (in \textit{light blue}) with the modified \textit{f-score} framework.
    (Note: the above figure represents a sample trajectory derived from synthetic points in order to preserve privacy).
    }
    \label{fig:missing_paths}
\end{figure}

In the IVMM algorithm \cite{yuan2010interactive}, the \textit{f-score} is an evaluation metric used to determine the optimal sequence of road network candidates for aligning GPS trajectories.

The proposed modification to the standard \textit{f-score} computation successfully addresses cases where missing paths in the road network prevent consecutive candidates from being connected. 
By adapting the \textit{f-score} assignment to split the weighted candidate graph into reachable subgraphs, the algorithm remains functional even when direct paths between certain pings are unavailable.
In scenarios where OpenStreetMap (OSM) data incompleteness leads to disconnected candidate pairs, the modified approach enables the Viterbi algorithm to continue processing without failure. 

In Fig \ref{fig:missing_paths} we demonstrate the effectiveness of our modified \textit{f-score} framework in addressing missing paths within a road network. 
Panel A illustrates the matched path \textit{red} obtained using the original IVMM approach, where trajectory points (in \textit{light blue}) are associated with their best candidates (in \textit{yellow}). 

However, due to missing road pieces in the road network (Fig \ref{fig:missing_paths} B \textit{green} area), the original method associate each ping with the nearest candidante, leading to an incorrect matching and to an improbable enhanced trajectory.

In contrast, Fig \ref{fig:missing_paths} C shows the matched path produced by our modified \textit{f-score} framework.
By splitting the weighted candidate graph into reachable subgraphs, our approach successfully connects trajectory points even in the presence of missing road segments. 
This adaptation ensures that the Viterbi algorithm remains robust and functional, enabling it to process trajectories effectively despite OSM data incompleteness. 
As a result, the matched path in panel C is correct compared to the one in panel A, underscoring the importance of our proposed modification in handling real-world scenarios where road networks may be incomplete or fragmented.
\section{Discussion}\label{sec12}

In this work, we have introduced several key enhancements to the original Interactive Voting-Based Map Matching (IVMM) algorithm, addressing both computational efficiency and robustness in the presence of heterogeneous GPS data.
Our distance-bounded voting mechanism has significantly reduced the quadratic computational complexity inherent in the traditional approach.
By limiting the interactions between distant pings using a well-chosen distance threshold, the algorithm scales more effectively to large datasets while maintaining its matching accuracy.

The improved treatment of missing road segments further contributes to the robustness of our method. 
In real-world scenarios, where map data (e.g., from OpenStreetMap) may be incomplete or fragmented, our modified \textit{f-score} framework ensures that optimal candidate sequences remain determinable by partitioning the candidate graph into reachable subgraphs. 
This adaptation is critical for applications in settings characterized by irregular or sparse mapping data.

Moreover, the integration of a custom-built road network asset derived from OpenStreetMap broadens the algorithm's applicability across different geographical regions. 
While the current work demonstrates significant improvements, several limitations remain that open up promising opportunities for future research.

First, the quality of the input GPS data is a critical factor that can significantly influence performance. 
In this regard, the integration of advanced pre-processing steps could further enhance data quality by reducing noise and mitigating the impact of irregular sampling rates \cite{huang2023accurate}.

Another limitation relates to the tuning of the spatio-temporal analysis function parameters. 
Our current implementation uses fixed values for these parameters; however, a data-driven parametrization may improve model performances.

A further challenge arises from the absence of comprehensive ground truth data. 
The evaluation of our approach currently relies on descriptive metrics—such as candidate-ping distance and path length variation—rather than direct comparisons against verified trajectories.
Our future work should focus on integrating ground truth datasets, either by crafting partially hidden datasets or by leveraging open-data repositories that provide trajectory data with marked sequences of roads traveled or even by generating ground truth using dedicated Software Development Kits (SDKs).
Access to such validated data would enable more rigorous performance assessments and facilitate targeted improvements in the matching process.

Finally, further exploration could focus on developing a method for assessing which trajectories are suitable for reconstruction using the approach we have proposed. 
Determining whether or not trajectory inference is feasible is not straightforward, as it depends not only on the spatial distance between GPS points but also on contextual factors such as the road network or geographical area. 
For example, if two GPS points on the same device are far apart but are located on a motorway, trajectory inference may be justified, as there may be only one plausible route connecting them.
Conversely, in an urban environment with multiple possible paths, trajectory reconstruction in these cases may not make sense.
A potential approach to distinguish between these scenarios is to count the number of distinct nodes in the road network between the two points and consider this information together with the spatial and temporal distance. 
If properly combined, these three variables could provide a systematic way to determine the cases in which reconstruction is justified.

Overall, addressing these limitations through enhanced pre-processing, adaptive parameter tuning, the integration of ground truth data, and the development of a methodology to determine when trajectory reconstruction is appropriate represents a promising direction for future research.
These efforts would not only strengthen the empirical basis of the map-matching approach, but also extend its usefulness in increasingly complex and heterogeneous data environments.

In summary, the enhanced IVMM algorithm represents a significant step forward in the map-matching domain. 
By effectively balancing computational efficiency with robustness against incomplete road network data, our approach offers a scalable solution for the challenges posed by low-sampling-rate and heterogeneous GPS trajectories. 
This work lays the foundation for further innovations in map matching and advanced trajectory analytics.

\section{Methods}\label{sec_methods}

\subsection{Data Description}

This study employs two distinct sources of anonymized and privacy-enhanced GPS trajectory data collected from connected vehicles and mobile phones, enabling us to evaluate the map-matching algorithm under varying quality conditions.

The connected car data, characterized by high-frequency recordings and precise positional accuracy, serves as an ideal benchmark for initial validation. 
Conversely, the mobile phone data reflects real-world variability through its comparatively sparse, irregular, and noisy readings.

By combining these sources, we thoroughly assess the algorithm’s adaptability: the vehicle data establishes a controlled baseline, while the mobile data tests resilience against sampling inconsistencies and positional uncertainty. 
This integrated approach ensures robustness for applications ranging from high-fidelity vehicular telematics to consumer-grade mobile analytics.

The following section provides a detailed description of each dataset.

\subsubsection{Connected Cars Data}

\begin{figure}[t!]
    \centering
    \includegraphics[width=\linewidth]{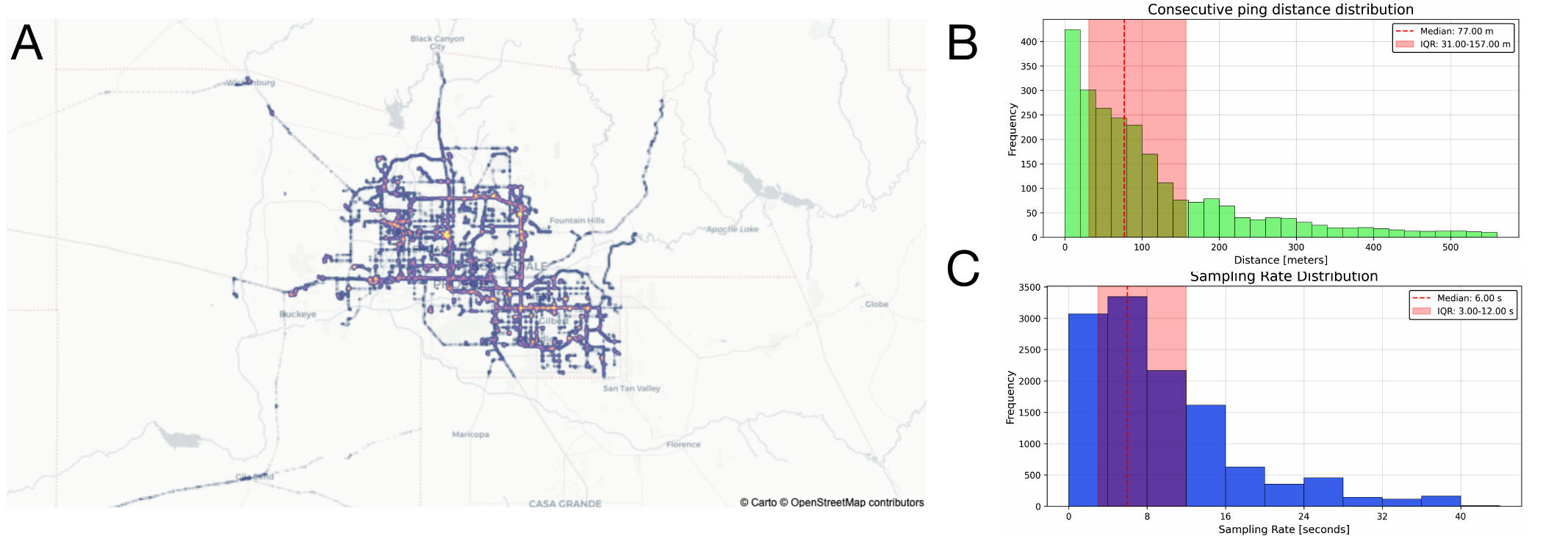} 
    \caption{\textbf{Connected cars data}. \textbf{A)} Spatial distribution of the GPS points data obtained from the connected vehicles. 
	\textbf{B)} Distribution of the distances in meters between GPS points of the same trajectory. 
    \textbf{C)} Distribution of the sampling rate. IQR stands for interquartile range.}
    \label{fig:connected_cars_data}
\end{figure}

This study used GPS location data from 203 connected cars in Maricopa County, Arizona, on May 30, 2023 provided by Wejo Inc (Fig \ref{fig:connected_cars_data} A). 
The dataset encompasses 500 anonymised trips, captured via onboard GPS systems with a high-frequency sampling rate of one point every 2-4 seconds (Fig \ref{fig:connected_cars_data} B).
Each data point includes geographical coordinates, timestamp, speed, bearing angle, event type (key on/off or mid-journey), journey ID, and vehicle type ID (make, model, and production year).
The dataset's high quality is evident in its detailed representation of vehicle movements. The inclusion of ignition status (KEY\_ON, KEY\_OFF, or MID\_JOURNEY) facilitates precise trajectory identification.

\subsubsection{Mobile Phone Data}

\begin{figure}[t!]
    \centering
    \includegraphics[width=\linewidth]{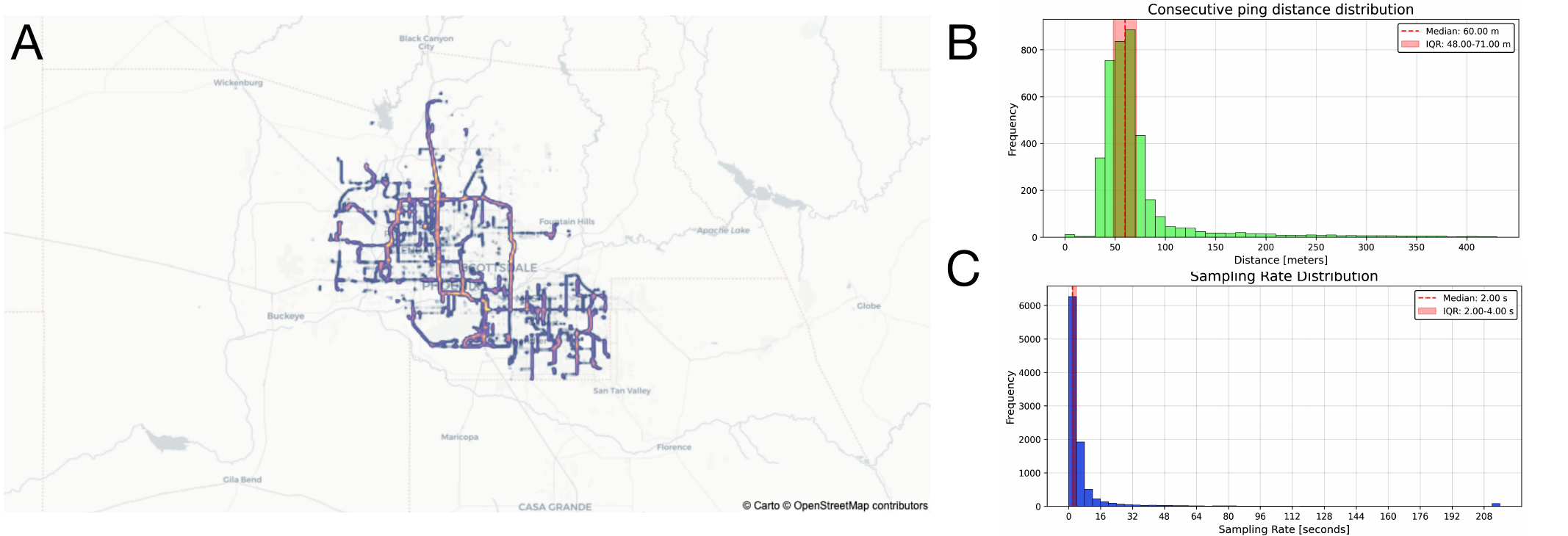} 
    \caption{\textbf{Mobile SDK data}. \textbf{A)} Spatial distribution of the GPS points data obtained from the mobile phone data. 
	\textbf{B)} Distribution of the distances in meters between GPS points of the same trajectory. 
    \textbf{C)} Distribution of the sampling rate. IQR stands for interquartile range.}
    \label{fig:devices_data}
\end{figure}

Complementing the vehicle data, this study also employed Cuebiq's proprietary dataset which contains anonymized GPS locations from opted-in mobile devices, for the same date and location (Fig \ref{fig:devices_data} A). 
This dataset comprises location data from 194 anonymous devices, representing 446 trajectories. 
Data collection adhered to GDPR and CCPA compliance, with users voluntarily opting to share their information for analytical and research purposes. 
In addition to anonymizing the data, Cuebiq applies a patented privacy-enhancing technology to obfuscate home locations in the data set to the Census Block Group level. Moreover, Cuebiq removes location records associated with privacy-sensitive points of interest. 
Cuebiq's Software Development Kit (SDK) gathered location data via GPS and Wi-Fi signals from both Android and iOS devices. 
Each data point is characterised by a unique device identifier, geographical coordinates (latitude and longitude), timestamp, speed, bearing angle, and position accuracy.
While GPS-enabled smartphones typically achieve accuracy within a 4.9-meter radius under optimal conditions, various factors can affect positioning accuracy.
These include signal blockage from structures or vegetation, indoor or underground usage, multipath effects, radio interference, solar storms, and satellite maintenance.
In Fig \ref{fig:devices_data} A and B are highlighted respectively the consecutive ping distance and the sampling rate distribution for this dataset.
To identify trajectories, the data were preprocessed using a density-based method for stop identification.
Trajectories were then defined as the sequence of GPS signals from a single user between consecutive stops.

\subsubsection{Road Network Data}
Our study employed a custom-built asset directly derived from OpenStreetMap (OSM) data \citep{OpenStreetMap}, specifically designed to support efficient and accurate ping-map matching.
OpenStreetMap is a collaborative, open-source geographic information system that provides free, editable map data of the world.
Founded in 2004, OSM has grown into a comprehensive global mapping platform maintained by a diverse community of volunteers and organisations, with data released under the Open Data Commons Open Database License (ODbL).
The OSM database contains a wide array of geographic features, including roads, buildings, natural landmarks, points of interest, and administrative boundaries.
Each element can have multiple tags, which are key-value pairs providing descriptive information about the feature.

Our custom asset was constructed by parsing the OSM planet data, filtering for \texttt{way\_id} that have the \texttt{highway} tag to capture all types of roads from motorways to mountain trails.
The \texttt{way\_id} in OpenStreetMap is a unique identifier for a "way", which represents a linear geographic feature like a road, river, or administrative boundary. 
A \texttt{way\_id} is defined as an ordered list of nodes (points with coordinates) that together form a path or a polygon.

Then, to provide more granular spatial information, we broke down each way into its rectilinear constituent, called road pieces.
Each road piece, defined by consecutive nodes of a way, was represented as a separate element in the asset, linked to its parent way via a \texttt{way\_id}.
Each road piece was enriched with the full original tags dictionary associated with that way, maintaining all the detailed information provided by OSM contributors.
In addition to the complete tags, we pre-extracted several key features to facilitate quick access and analysis: \texttt{highway} (indicating the type of road), \texttt{maxspeed} (specifying the maximum legal speed limit where available, standardised in km/h) and \texttt{oneway} (a boolean flag indicating one-way streets), \texttt{service} (describing special types of service roads).

For this study, we focused on two key variables from the OSM road network data: \texttt{maxspeed} and \texttt{oneway}.
These variables were crucial for implementing a weighted directed graph for the road network, where weights were given by the speed constraints of the road pieces and the directionality of the links was taken into account.
Analysis of the OSM road network data revealed significant challenges, with 94.53\% of missing data in the \texttt{maxspeed} field for US roadways and a high number of null values in the \texttt{oneway} field.

To address the missing \texttt{maxspeed} data, we employed an imputation method.
We computed mean values per road type per admin1 level, taking into account different legislation for velocity constraints in different states.

For the \texttt{oneway} field, unlike for \texttt{maxspeed} where simple imputation is a viable solution, we considered two potential approaches:

\begin{enumerate}
    \item Applying a Graph Neural Network (GNN) model \cite{kipf2016semi, hamilton2017inductive, velivckovic2018graph} to leverage the adjacency structure of the road network along with other OSM attributes to predict the \texttt{oneway} field \cite{he2020roadtagger}.
    \item Applying default values where explicit OSM tagging exists (e.g., \texttt{highway=motorway} or \texttt{highway=motorway\_link} implies \texttt{oneway=True}); otherwise, assuming the road to be two-way (\texttt{oneway=False}).
\end{enumerate}

While the GNN-based approach is innovative and data-driven, the default rule-based method guarantees the addition of correct directional links to the network, so we decided to use the latter.

To assess the missing data, we analyzed the distribution of known \texttt{oneway} values: 79.2\% were \texttt{oneway=True}, with motorways, link roads, and bus lanes exhibiting the highest prevalence of one-way classification.
However, other road types showed no clear pattern. 
Some anomalies, such as motorways tagged as \texttt{oneway=False}, were traced to temporary construction zones where traffic flow was modified. Additionally, certain footways tagged as \texttt{oneway=True} appeared to restrict only vehicular movement while allowing bidirectional pedestrian access.
For our GNN-based approach, we constructed a graph representation where nodes corresponded to road segments, and edges connected adjacent segments. The graph was built using rounded latitude and longitude coordinates of segment endpoints, ensuring connectivity while removing small, isolated road components (e.g., internal paths in closed areas like zoos or cemeteries). 
The final graph comprised 253,387 nodes.

Our GNN model used input features including \texttt{maxspeed}, length, number of lanes, and \texttt{highway} classification. The output was a binary classification: \texttt{oneway=True} or \texttt{oneway=False}. 
Due to severe class imbalance, we applied undersampling to create a balanced training set (50\% one-way, 50\% two-way). 
The model architecture included two GraphSAGE convolutional layers \cite{hamilton2017inductive} (each with 32 hidden units), trained with a batch size of 64, learning rate of 0.01, and dropout rate of 0.005.
After 100 epochs, the model achieved an overall validation accuracy of ~81\%, with precision and recall of ~0.83. 

However, despite these promising results, continuity issues emerged: the predicted \texttt{oneway} classifications lacked consistency along individual roads, leading to segments within the same street being intermittently classified as one-way and two-way.
This suggests that additional smoothing techniques or probabilistic classification thresholds may be required to enhance spatial coherence.
In the end, we opted for the default value approach, which proved to be consistent and accurate even through sample analysis.

\begin{table}[t!]
\centering
\renewcommand{\arraystretch}{1.2}
\begin{tabular}{rl}
\toprule
\textbf{Notation} & \textbf{Description} \\
\midrule
$p_i$ & GPS ping at index $i$ \\
$c_i^j$ & Candidate $j$ for GPS ping $i$ \\
$e_u$ & Road piece $u$ \\
$x_i^j$ & Distance between ping $i$ and candidate $j$ \\
$d_{i-1 \rightarrow i}$ & Physical distance between two consecutive pings $i-1$ and $i$ \\
$w_{(i-1,t) \rightarrow (i,s)}$ & Length of the shortest path between candidate $t$ of ping $i-1$ and candidate $s$ of ping $i$ \\
$\bar{v}_{(i-1,t) \rightarrow (i,s)}$ & Average speed of the shortest path between two candidates \\
$\Delta t_{i-1 \rightarrow i}$ & Time interval between two sampling points $p_i$ and $p_{i-1}$ \\
$e_u' \cdot v$ & Typical speed value associated with road piece $e_u'$ \\
$maxdist$ & Distance threshold for the distance-bounded interactive voting implementation\\
$n$ & Number of pings in a trajectory \\
$\mu$ & Mean of the Gaussian distribution used as \textit{observation probability} \\
$\sigma$ & Standard deviation of the Gaussian distribution used as \textit{observation probability} \\
$g(x)$ & Distance function used in the interactive voting process \\
$\beta$ & Distance weight function coefficient with respect to the road network \\
$\alpha$ & Buffer radius for spatial join between road pieces and trajectory pings \\
$k$ & Maximum number of candidates retained for each ping \\
$minpings$ & Minimum number of pings required for a trajectory \\
\bottomrule
\end{tabular}
\caption{Notation and parameters used with their description}
\label{tab:notation}
\end{table}

\subsection{Algorithm Description and Enhancements}\label{algo_description}

\begin{figure}[ht!]
    \centering
    \includegraphics[width=\linewidth]{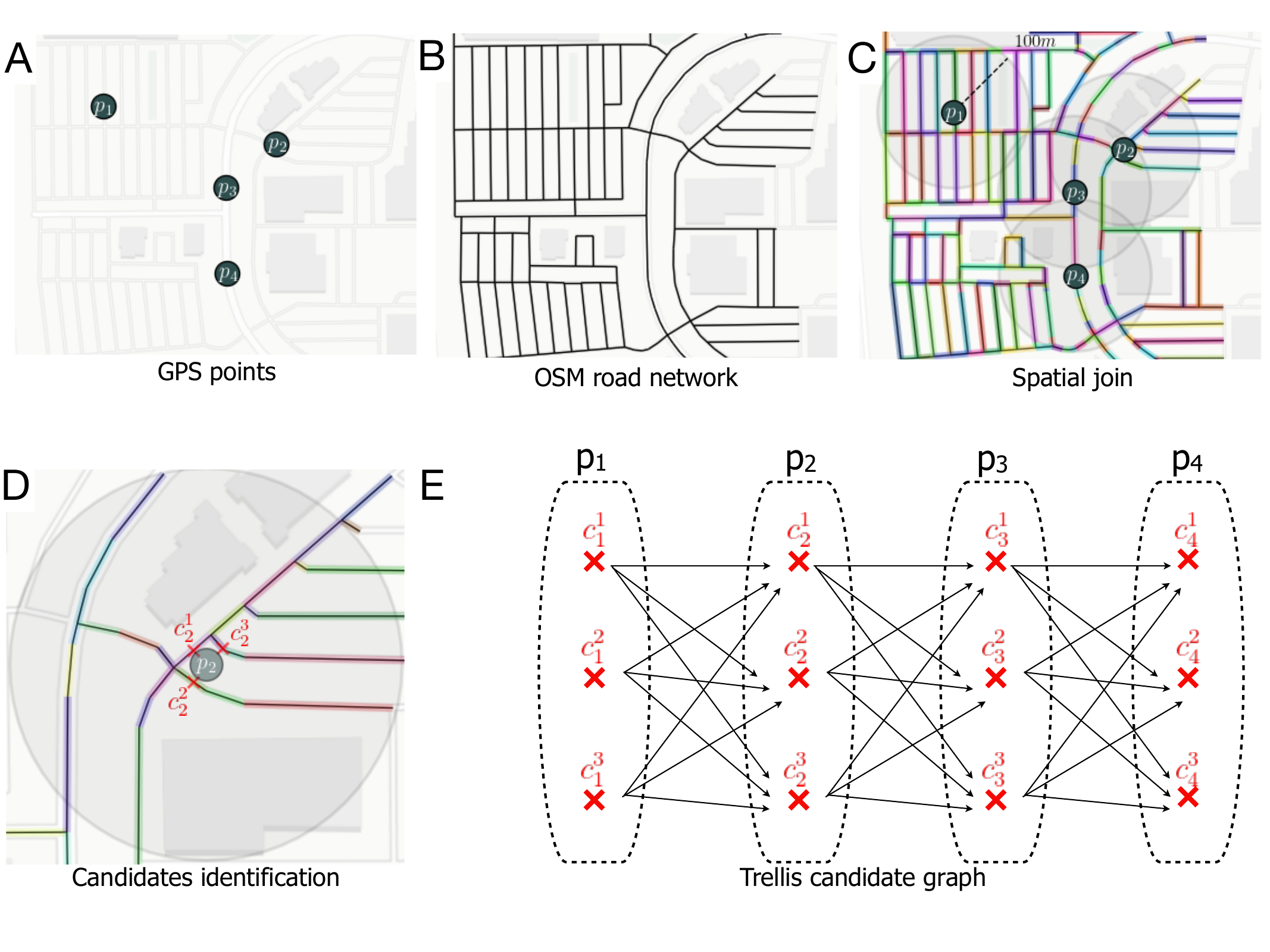} 
    \caption{\textbf{From trajectory to Trellis graph}. 
    \textbf{A)} An example of a simple trajectory consisting of 4 points numbered from first to last in order of time.
    \textbf{B)} The road network obtained via OSM in that geographic area.
    \textbf{C)} The spatial join between the series of GPS points and the road pieces (each road piece has a different color) obtained via OSM. The buffer around each GPS point provides the area within which candidates will be searched.
    \textbf{D)} An example of the candidate identification for point 2.
    \textbf{E)} The Trellis graph for the trajectory considered.
    }
    \label{fig:algo description}
\end{figure}

In this section, we describe the operation of the algorithm, highlighting the contributions and enhancements introduced in this work.
Our methodology for ping-map matching builds upon the work of \citet{yuan2010interactive}, with enhancements to improve efficiency and accuracy.
To better clarify the formal terminology of the following description, a notation dictionary is provided in Table \ref{tab:notation}.

Starting with a trajectory of GPS points—representing a sequence of temporally ordered locations (as showed in Fig. \ref{fig:algo description} A), the candidates were identified via spatial join between road pieces and trajectory pings within a buffer of radius $\alpha$ (Fig. \ref{fig:algo description} C).
The resulting candidates are projections of GPS points onto nearby road pieces that fall within this buffer (Fig. \ref{fig:algo description} D). 
To manage computational complexity, only the $k$ candidates with the smallest physical distance from each ping were retained.

The initial candidate graph has been constructed in the form of a Trellis graph \citep{ryan1993viterbi} (Fig. \ref{fig:algo description} E), where nodes represent candidates grouped in slices for each ping, and links connect pairs of candidates belonging to subsequent pings.
This graph is directed, with links oriented towards subsequent pings, and weighted.
The link weights are obtained as the product of three quantities: 
\begin{itemize}
    \item the observation probability, which represents the distribution of the measurement error of the GPS signal; 
    \item the transition probability between candidates, represented as the ratio of physical distance to shortest path length; 
    \item the cosine distance between sample and typical road speed. 
\end{itemize}

This methodology used by Yuan et al. \cite{yuan2010interactive} was originally developed and introduced by Lou et al. \cite{lou2009map}.\\
The \textbf{observation probability} of the candidate $c_i^j$ is:

\begin{equation}
    N(c_i^j) = \frac{1}{\sqrt{2\pi\sigma}} e^{-\frac{(x_i^j - \mu)^2}{2\sigma^2}}
\end{equation}

where $x_i^j$ is the distance between ping $i$ and candidate $j$, and $\mu$ and $\sigma$ are parameters of the Gaussian distribution (Table \ref{tab:notation}). 
The choice of a Gaussian distribution is justified by the assumption that $x_i^j$ represents an accuracy error, typically modeled as white noise around the true measurement. 
The parameter $\sigma$ is related to the expected GPS accuracy, while $\mu$ is often set to 0, although it may deviate from 0 to account for potential systematic errors.

Here, we introduce the first modification made to adapt the algorithm for OSMs road network.
Specifically, the \textbf{transition probability} between candidates has been slightly adjusted from the formulation presented in \cite{yuan2010interactive}.

In the original work, the transition probability was defined as the ratio of the Euclidean distance between two successive pings to the length of the shortest path between the candidates corresponding to these pings. 
As described earlier, candidates are projections of GPS points onto nearby road segments. 

However, in certain cases—particularly at road junctions—the shortest path between candidates may be shorter than the Euclidean distance between successive pings.  
To generalize the formula and ensure the transition probability always remains within a valid probabilistic range, we redefine it as follows:  

\begin{equation}
V(c_{i-1}^t \rightarrow c_i^s) = \frac{\min(d_{i-1\rightarrow i}, w_{(i-1,t)\rightarrow(i,s)})}{\max(d_{i-1\rightarrow i}, w_{(i-1,t)\rightarrow(i,s)})}
\end{equation}  

Here, the transition probability is computed as the ratio of the smaller value to the larger value between: 

\begin{itemize}
    \item the physical distance \( d \) between two consecutive pings \( i-1 \) and \( i \), and  
    \item the length of the shortest path \( w \) between candidate \( t \) of ping \( i-1 \) and candidate \( s \) of ping \( i \). 
\end{itemize} 

The \textbf{temporal weight function}, which compares the average sample speed $\bar{v}$ to the typical road speed $v$, is computed using the cosine distance between these two quantities (as in the original methodology).

In our framework as the typical speed of the road piece we used the value of the field \texttt{maxspeed} obtained from the OSM road network.

Given two candidate points $c_{i-1}^t$ and $c_i^s$ for two neighboring GPS sampling points $p_{i-1}$ and $p_i$ respectively, the shortest path from $c_{i-1}^t$ to $c_i^s$ is denoted as a list of road pieces $[e_1', e_2', \ldots, e_k']$. 
The average speed $\bar{v}_{(i-1,t)\rightarrow(i,s)}$ of the shortest path is computed as follows:

\begin{equation}
\bar{v}_{(i-1,t)\rightarrow(i,s)} = \frac{\sum_{u=1}^k l_u}{\Delta t_{i-1 \rightarrow i}}
\end{equation}

where $l_u = e_u'l$ is the length of $e_u'$, and $\Delta t_{i-1 \rightarrow i}$ is the time interval between two sampling points $p_i$ and $p_{i-1}$.
Note that each road piece $e_u'$ is also associated with a typical speed value $e_u' v$.

The cosine distance is used to measure the similarity between the actual average speed from $c_{i-1}^t$ to $c_i^s$ and the speed constraints of the path.

The temporal weight function is then defined as:

\begin{equation}
F_t(c_{i-1}^t \rightarrow c_i^s) = \frac{\sum_{u=1}^k (e_u' \cdot v \times \bar{v}_{(i-1,t)\rightarrow(i,s)})}{\sqrt{\sum_{u=1}^k (e_u' \cdot v)^2} \times \sqrt{\sum_{u=1}^k \bar{v}_{(i-1,t)\rightarrow(i,s)}^2}}
\end{equation}

As in the spatial analysis function, $c_{i-1}^t$ and $c_i^s$ are any two candidate points for $p_{i-1}$ and $p_i$ respectively.
The product of the three quantities defined above yields the \textit{ST-Matching} function, which assigns weights to the links in the Trellis graph.

The final step of the algorithm is the interactive voting process, which we retain as presented in \citep{yuan2010interactive}.

For each ping, the weights of the candidate graph are modified, and the optimal sequence for each candidate is determined. 
Initially, the same candidate graph is assigned to every ping. 
For each iteration, a modified version of this graph is generated based on the current reference ping.

The weights of the candidate graph are adjusted to incorporate the physical distance between the reference ping and the remaining pings.
This distance serves as a measure of the interaction strength between pings, with greater distances resulting in weaker connections between candidates.
This is obtained using a suitable distance function $g(x)$. 
Any decreasing function satisfying $g(0)=1$, $g(\infty)=0$, with values in the interval $[0,1]$ is appropriate.
As in \citet{yuan2010interactive} we used for the distance function:
\begin{equation}
    g(x) = e^{-\frac{x^2}{\beta^2}}
\end{equation}

For each candidate of the reference ping, we identify the optimal path through the graph, constrained to pass through that specific candidate.
This optimal path is determined by maximising the sum of graph weights, which we define as the \textit{f-score}.
To compute the \textit{f-score} and find the optimal sequence, we employ the Viterbi algorithm.
This process yields one optimal sequence per candidate for each reference ping.\\
The voting mechanism operates as follows:
\begin{enumerate}
    \item Each candidate that appears in an optimal sequence receives a vote.
    \item The total votes for each candidate are tallied and updated with each iteration.
    \item In addition to votes, each candidate is assigned a score equal to the \textit{f-score} of the optimal path it belongs to.
    \item The total score for each candidate is also updated per iteration.
\end{enumerate}

This dual scoring system—combining vote counts and path scores—provides a robust method for evaluating the likelihood of each candidate being part of the true trajectory.
By considering both the frequency of a candidate's appearance in optimal paths (votes) and the quality of those paths (scores), we can more accurately determine the most probable map-matched route.

\subsubsection{Distance-Bounded Interactive Voting Implementation}
The final phase of the original map matching method, as described by \citet{yuan2010interactive}, is the most time consuming.
Each candidate for every ping votes for the optimal candidate of all other pings. 
The computational complexity of this procedure reaches $O(n^2k^2)$, where $n$ represents the number of pings and $k$ denotes the maximum number of candidates per ping.
This can result in exceptionally high computational times, particularly for long trajectories, thus limiting the feasibility of extensive training, especially on local machines.

To address these computational challenges, we developed a methodology that implements a distance-bounded version of the interactive voting process.
This approach limits the voting process to pings within a meaningful distance, more realistically reflecting the diminishing influence of distant pings.

The process can be described as follows:

\begin{enumerate}
    \item For each reference ping, we compute distances to all other pings in the trajectory.
    \item We apply a distance threshold, removing ping slices where the distance to the reference ping exceeds this predefined threshold.
    \item We compute the optimal sequence for each candidate of the reference ping using the Viterbi algorithm on the reduced candidate graph.
\end{enumerate}

From a modeling perspective, this distance-bounded implementation is equivalent to using a distance function with a cutoff at a predefined threshold.

Mathematically, given a threshold $maxdist$, the modified distance function $g'(x)$ can be expressed as:

\begin{equation}
    g'(x) = 
    \begin{cases}
        g(x) & \text{if } x < maxdist \\
        0 & \text{otherwise}
    \end{cases}
\end{equation}

where $g(x)$ is the original distance function.

To analyse the trade-off between computational efficiency and matching accuracy, we tested multiple buffer sizes: 1000, 2500, 4000, and 5500 meters. 
This allowed us to determine an optimal buffer size that balances processing speed with map matching precision.

\subsubsection{Handling Missing Paths}
During the construction of the candidate graph, the \textit{ST-Matching} function computes the length of the shortest path on the road network graph, which is then incorporated into the edge weights. 
For any pair of candidates $(c_i, c_j)$, if there is no path between them in the road network (which can occur due to incompleteness in the OSM data), an infinite distance is assigned.
Consequently, the edge connecting these candidates is removed from the candidate graph by setting its weight to $-\infty$. 

A problem arises when two consecutive pings have no existing connections between any of their candidates.
In such cases, all edges of the Trellis graph between two consecutive slices are removed (set to $-\infty$), preventing the optimal sequence search using the Viterbi algorithm.

To address situations where no paths exist on the road graph between the candidates of two consecutive pings, we implemented a modified \textit{f-score} computation process.
This process handles missing paths while maintaining the integrity of the standard case by effectively dividing the weighted candidate graph into subgraphs where all pings can reach their consecutive neighbors.

The modified \textit{f-score} assignment process follows this strategy:

\begin{enumerate}
    \item For the candidates of the first ping, the \textit{f-score} is set to their \textit{ST-Matching} function value, weighted by the distance from the reference ping.
    \item For candidates of subsequent pings (from the second to the last), the \textit{f-score} is determined as follows:
    \begin{itemize}
        \item If a valid path exists, it is set as the maximum of the sums of the previous candidate's \textit{f-score} and the weight of the edge from the previous to that candidate (as in the standard case).
        \item If no valid path exists (i.e., the maximum is $-\infty$), it is set equal to the \textit{ST-Matching} function value weighted by the distance from the reference ping (similar to the first ping in the standard case).
    \end{itemize}
\end{enumerate}

Intuitively, this approach corresponds to splitting the weighted candidate graph into subgraphs where all pings can reach their consecutive neighbors.
The total \textit{f-score} of the procedure can thus be computed as the sum of the total \textit{f-scores} in each candidate subgraph.

This modification allows the algorithm to handle incomplete road network data more robustly, maintaining the ability to find optimal paths even when direct connections between some consecutive pings are missing.
It preserves the integrity of the map-matching process while adapting to real-world imperfections in geographic data.

\bibliography{references}

\end{document}